\documentclass[nohyperref]{article}

\usepackage{microtype}
\usepackage{graphicx}
\usepackage{subfigure}
\usepackage{booktabs} 
\usepackage{mathtools}
\usepackage{multirow}
\usepackage{hyperref}
\usepackage{float}
\usepackage{placeins}

\usepackage[accepted]{icml2022}

\usepackage{amsmath}
\usepackage{amssymb}
\usepackage{mathtools}
\usepackage{amsthm}

\usepackage[capitalize,noabbrev]{cleveref}

\theoremstyle{plain}

\theoremstyle{definition}

\theoremstyle{remark}

\usepackage[textsize=tiny]{todonotes}

\icmltitlerunning{The Importance of Background Information for Out of Distribution Generalization}

\begin{document}

\twocolumn[
\icmltitle{The Importance of Background Information for Out of Distribution Generalization}




\begin{icmlauthorlist}
\icmlauthor{Jupinder Parmar}{cs}
\icmlauthor{Khaled Saab}{ee}
\icmlauthor{Brian Pogatchnik}{rad}
\icmlauthor{Daniel Rubin}{bme}
\icmlauthor{Christopher Ré}{cs}
\end{icmlauthorlist}

\icmlaffiliation{cs}{Department of Computer Science, Stanford University, Stanford, USA}
\icmlaffiliation{ee}{Department of Electrical Engineering, Stanford University, Stanford, USA}
\icmlaffiliation{bme}{Department of Biomedical Data Science, Stanford University, Stanford, USA}
\icmlaffiliation{rad}{Department of Radiology, Stanford University, Stanford, USA}

\icmlcorrespondingauthor{Jupinder Parmar}{jsparmar@stanford.edu}

\icmlkeywords{Machine Learning, Domain Generalization, Robustness, ICML}

\vskip 0.3in
]



\printAffiliationsAndNotice{} 

\begin{abstract}

Domain generalization in medical image classification is an important problem for trustworthy machine learning to be deployed in healthcare. We find that existing approaches for domain generalization which utilize ground-truth abnormality segmentations to control feature attributions have poor out-of-distribution (OOD) performance relative to the standard baseline of empirical risk minimization (ERM). We investigate what regions of an image are important for medical image classification and show that parts of the background, that which is not contained in the abnormality segmentation, provides helpful signal. We then develop a new task-specific mask which covers all relevant regions. Utilizing this new segmentation mask significantly improves the performance of the existing methods on the OOD test sets. To obtain better generalization results than ERM, we find it necessary to scale up the training data size in addition to the usage of these task-specific masks. 

\end{abstract}

\section{Introduction}

Machine learning models have been shown to be extremely brittle to distributional shifts from the training set \cite{Geirhos2020}. This causes models to fail when evaluated on out-of-distribution (OOD) test sets. Oftentimes, the presence of spurious correlations: misleading heuristics that are correlated with the label in the training data yet are independent of the label in the target domain, are an underlying reason for this phenomenon \cite{chen2020, parascandolo2020, hermann2020}. For example, on the task of identifying pneumonia in chest X-rays, confounding factors like the presence of hospital-specific tokens in radiographs cause models to fail when given scans from new hospitals \cite{Zech2018}. Hence, it is pertinent to understand how methods that are more robust to distributional shifts can be developed. 


Through our experiments, we find that methods which regularize models to only focus on abnormal regions \cite{Ross17, Viviano2019} do not perform better than empirical risk minimization (ERM) for medical image classification. Although deriving features in this manner reduces reliance on spurious correlations \cite{pmlr-v102-zhuang19a}, these methods still result in poor generalization. 

We hypothesize that select parts of the image background, regions not contained in the segmentation map, provide generalizable signal for the task and that one of the reasons current approaches fail is due to these regions being left out of the feature attribution. For instance, when given an image it has been shown that humans do not solely fixate on the focal point (the location of the pathology), but contextualize by viewing the surrounding area \cite{kirtley2018}.

In this work, we focus on the task of identifying pneumothorax (i.e, a collapsed lung) within chest X-rays. Our domain generalization setting has one source domain and four manually curated target domains, based on shifts of age of population or hospital of collection. To identify which regions of the image background are relevant, we first analyze gaze data obtained from domain-experts for the task on the source domain \cite{saab2021}. Human gaze information has been shown to embed information about the regions a viewer considers to be important \cite{Yun2013}, hence it can be used to find areas that the experts deem to contain signal for the task. From this analysis, we develop a segmentation map of the lung periphery as we find that it includes all relevant regions for the task.

Utilizing these task-specific segmentation maps, our methods obtain an increase of 4.43 points in AUROC on average over the 4 target domains in comparison to using ground-truth segmentation masks. Additionally, the new masks help the methods beat ERM by an average of 1.35 points in AUROC across the 4 target domains. Our findings suggest that when chosen correctly, periphery data outside the region of abnormality \emph{can} improve generalization to OOD samples.

\section{Related Work}


To constrain learned features towards areas of interest, one avenue in domain generalization literature has been on directly controlling learned attributions via saliency gradients. By penalizing saliency gradients that appeared outside of the ground-truth segmentation region of an image, Right for the Right Reasons (RRR) was the first to show improved generalization performance on a synthetic dataset \cite{Ross17}. More recent methods such as GradMask \cite{Simpson2019} have built on RRR to realize small improvements on real data. 

Alternatively, another family of approaches that seek to learn high-level features and has been effective on domain generalization is domain invariant representation learning \cite{gulrajani2020, koh2020}. These methods aim to output latent features of the model that are indistinguishable across domains. Approaches include utilizing an adversarial network to identify a domain \cite{ganin2015} and matching domains through a contrastive loss \cite{motiian2017}. More recently, ActDiff \cite{Viviano2019} has found promising results on synthetic and real data by finding features that are invariant to regions outside of the ground-truth segmentation of an image.




\section{Problem Setup}

\textbf{Domain Generalization}

Let $\mathcal{X}$ and $\mathcal{Y}$ be the feature and label spaces, respectively. A domain is defined as a joint distribution over $\mathcal{X} \times \mathcal{Y}$. A learning model is defined as $f: \mathcal{X} \rightarrow \mathcal{Y}$. In our domain generalization study, there is a single source domain $\mathcal{D}_S$ and $L$ target domains $\{\mathcal{D}_i\}_{i=1}^L$, where $L = 4$. The samples available to us at training time are $S_{src}$ which are taken as i.i.d samples from $\mathcal{D}_S$ . The goal of domain generalization is to learn a model $f$ using data from the source samples such that the model can generalize well to samples from the unseen target domains at test time.

\textbf{Datasets}

We focus on the binary classification task of pneumothorax identification within chest radiographs. We specify our source and target datasets below. All datasets contain the same class balance. 

\textbf{CXR-P (Source)}: We use the CXR-P dataset introduced in \cite{saab2021} as our source dataset. It consists of 5,777 X-ray images of which 22\% contain pneumothorax. 1,170 images are reserved for the train and validation sets with the remaining 4,607 images forming the held-out test set. CXR-P was originally sourced from the SIIM-ACR Pneumothorax dataset \cite{society2010pneumothorax} which consists of 10,675 chest radiographs with ground-truth segmentation maps for abnormal images. 

\textbf{ChestX-ray8 (Target)} ChestX-ray8 is a dataset of chest radiographs collected from the NIH Clinical Center hospital. SIIM-ACR is a subset of ChestX-ray8 and hence both are collected from the same hospital. Thus, to curate an OOD evaluation set we focus on the distributional shift of age of population. We first remove all SIIM-ACR images from ChestX-ray8 to prevent data leakage and randomly sampled a subset of 4,607 individuals with ages above the median for the dataset. This newly defined OOD set has an average age of 71.39 compared to 51.12 for CXR-P.

\textbf{MIMIC-CXR (Target) } We sample 4,607 images from the MIMIC-CXR dataset \cite{johnson2019mimic} which contains chest X-rays sourced from the Beth Israel Deaconess Medical Center Emergency Department \textemdash \ a different hospital than CXR-P. As MIMIC-CXR does not release patient demographics, this evaluation set solely consists of a distribution shift in hospital of collection.


\textbf{CheXpert (Target) } Our third and fourth target datasets are sourced from CheXpert \cite{irvin2019chexpert}. The images in CheXpert are collected from the Stanford Hospital and patient demographics are also released allowing us to utilize both distributional shifts. We first sample a target set of 4,607 chest X-rays whose average patient age is 54.40 and then create a second target of 4,607 chest X-rays whose average patient age is 89.60 by using the same methodology outlined for ChestXray-8. Hence, the first dataset, which we title CheXpert, only has a distribution shift of hospital of collection while the second dataset, which we refer to as CheXpert-age, additionally has a shift of age of population.

\section{Domain Generalization Methods}

We study the performance of two methods for domain generalization that rely on ground-truth segmentation maps:

 \textbf{ActDiff} \cite{Viviano2019} first creates a masked version of each input, \textbf{x}, using the ground-truth segmentation map as follows: $\textbf{x}_{masked} = \textbf{x}\cdot \textbf{x}_{seg} + \text{shuffle}(\textbf{x}) \cdot (1 - \textbf{x}_{seg})$. The shuffle function randomly permutes values in the background of the image to remove any spatial information. ActDiff then optimizes:
 \begin{equation*}
\mathcal{L}_{act} = \smash{\sum_{\textbf{x}, \textbf{x}_{masked} \in D}} \mathcal{L}_{clf} + \lambda_{act} ||o_{l}(\textbf{x}_{masked} - o_{l}(\textbf{x})||_2
\end{equation*}

where $o_{l}(\cdot)$ are the pre-activation outputs for layer $l$ of the encoder $f(\cdot)$ and $\mathcal{L}_{clf}$ is the standard cross entropy loss.
  
\textbf{Right for the Right Reasons (RRR)} \cite{Ross17} pushes the saliency gradients of the summed log probabilities of the $K$ output classes to zero in regions outside of the ground-truth segmentation map by minimizing: 
  \begin{equation*}
\mathcal{L}_{rrr} = \smash{\sum_{\textbf{x}, \textbf{x}_{seg} \in D}} \mathcal{L}_{clf} + \lambda_{rrr} \left[(1 - \textbf{x}_{seg}) \cdot \frac{\partial }{\partial \textbf{x}}\sum_{k = 1}^{K} \log(\hat{p}_k)  \right]^2
\end{equation*}

We hypothesize that as these methods completely ignore the background when performing feature attribution, they will fail on generalization performance for pneumothorax classification. Given that medical imaging is a challenging real-world application, helpful discriminative features for the task likely are not constrained to just the abnormality segmentation. To evaluate the performance of these methods, we will look at their AUROC on the OOD target datasets in comparison to the standard baseline of ERM.  

We implement both methods, along with ERM, and use a ResNet-50 CNN \cite{He2015} from Torchvision \cite{Marcel2010TorchvisionTM} pretrained on ImageNet as the backbone of our learning model in all experiments. Hyperparameters were chosen by tuning over a grid for learning rate, weight decay, $\lambda_{actdiff}$, and $\lambda{rrr}$ with best found hyperaparameters specified in the Appendix. We trained ERM for 15 epochs per \cite{saab2021} while all other methods were trained for 100 epochs per \cite{Viviano2019} .

\FloatBarrier
\begin{table}[ht]
\caption{ Results are averaged over 10 random seeds with 95\% significance. The target datasets are underlined. Neither method exceeds ERM in OOD performance.
}
\label{initial_results}
\begin{center}
\begin{small}
\begin{sc}
\begin{tabular}{lccr}
\toprule
Dataset & Method & AUROC \\
\midrule
\multirow{3}{*}{CXR-P} & ERM    & \textbf{83.5} $\pm$ 1.2&  \\
& ActDiff & 78.6$\pm$ 2.2& \\
& RRR & \textbf{83.5} $\pm$ 1.4 & \\
\cline{0-3}
\multirow{3}{*}{\underline{MIMIC-CXR}} & ERM    & \textbf{77.5}$\pm$ 2.2&  \\
& ActDiff & 68.7$\pm$ 6.3& \\
& RRR & 76.6 $\pm$ 3.0 & \\
\cline{0-3}
\multirow{3}{*}{\underline{CheXpert}} & ERM    & \textbf{80.7}$\pm$ 1.5&  \\
& ActDiff & 71.6$\pm$ 5.6& \\
& RRR & 79.2 $\pm$ 2.0 & \\
\cline{0-3}
\multirow{3}{*}{\underline{CheXpert-age}} & ERM    & \textbf{74.0}$\pm$ 1.6  \\
& ActDiff & 68.1$\pm$ 4.4& \\
& RRR & 71.6 $\pm$ 2.3 & \\
\cline{0-3}
\multirow{3}{*}{\underline{ChestX-ray8}} & ERM    & \textbf{73.4}$\pm$ 1.1&  \\
& ActDiff & 69.8$\pm$ 1.3& \\
& RRR & 73.1 $\pm$ 1.6& \\
\bottomrule
\end{tabular}
\end{sc}
\end{small}
\end{center}
\end{table}
\FloatBarrier

\cref{initial_results} shows that both \textit{ActDiff} and \textit{RRR} have worse OOD performance than ERM. On average, ActDiff and RRR did 6.85 and 1.53 points worse than ERM on the target sets, respectively. These results validate that the features learned by these methods are not capturing the necessary information to successfully learn the task of pneumothorax classification and improve OOD performance

\section{Utility of Background Information}

To better learn robust and generalizable features, we hypothesize that more information on relevant background regions is required. Given that ground-truth segmentations cover the smallest region associated with an abnormality, expanding the scope of the feature attribution of these methods is likely key. To investigate this hypothesis, we make use of the released gaze information from \cite{saab2021} which consists of the image locations a domain expert's eyes fixated on during labeling time. Previous studies have shown that gaze data contains task-relevant information \cite{hayhoe2005}, hence by finding the most commonly visited regions in the gaze sequences we can determine areas of the radiograph that are deemed relevant by domain experts.

\FloatBarrier
\begin{figure}[h]
\begin{center}
\centerline{\includegraphics[scale=0.18]{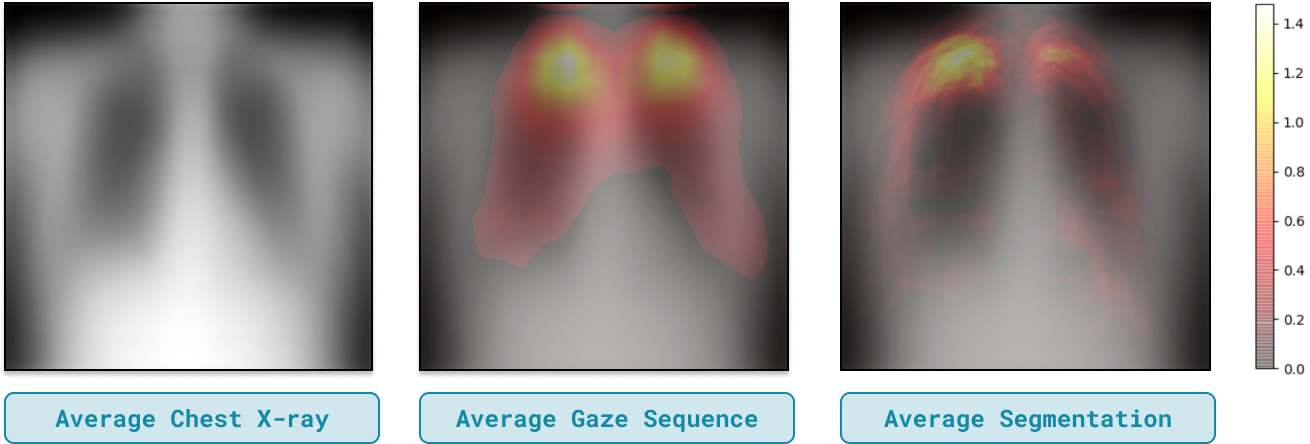}}
\caption{The average domain-expert gaze sequence focuses on the general region of the lungs. The average segmentation map shows that most pneumothorax is found in the periphery of the lungs.}
\label{overlays}
\end{center}
\end{figure}
\FloatBarrier

\FloatBarrier
\begin{table*}[h]
\caption{Test results for ActDiff when utilizing different segmentation maps. The lung periphery segmentations consistently win across the target sets. Target datasets are underlined. Results are averaged over 10 random seeds with 95\% significance. }
\label{lung_periphery_results}
\begin{center}
\begin{small}
\begin{sc}
\begin{tabular}{lcccccr}
\toprule
Segmentation Type & CXR-P & \underline{MIMIC-CXR} & \underline{CheXpert} & \underline{CheXpert-age} & \underline{ChestX-ray8} \\
\midrule
Abnormality & 78.6$\pm$ 2.2 & 68.7$\pm$ 6.3 & 71.6$\pm$ 5.6 & 68.1$\pm$ 4.4 & 69.8$\pm$ 1.3  \\ 
Scaled Abnormality & 79.6$\pm$ 2.2 & 73.8$\pm$ 2.8 & 73.3$\pm$ 3.9 & 68.6$\pm$ 2.5 & 71.1 $\pm$ 2.2  \\
Lung Periphery & \textbf{82.8} $\pm$ 0.9 & \textbf{77.6} $ \pm$ 2.1 & \textbf{80.2} $\pm$ 2.4 & \textbf{72.6} $\pm$ 2.1 &  \textbf{73.1} $\pm$ 1.5  \\
\bottomrule
\end{tabular}
\end{sc}
\end{small}
\end{center}
\end{table*}

The released gaze data exists for the training split of CXR-P. We look at the average gaze sequence across these images, filtering out low values which indicate infrequently visited positions, and overlay it on top of the average of all images in the set. \cref{overlays} demonstrates that the highest propensity of gaze fixations occur at the top of the lungs with the entire lung area, except the center, generally being taken into consideration. As there is noise in the fixations of the gaze data \cite{saab2021}, to validate that the identified regions are relevant we discuss the findings with a radiologist who verifies that the lungs, specifically the lung periphery, are the most relevant areas for pneumothorax classification.

To validate the radiologist's claim, \cref{overlays} shows the average abnormality segmentation map overlayed on the average image of the training split. This illustrates that while the largest number of abnormalities occur in the upper lungs, the entire periphery is needed to capture the vast majority of segmentations. Thus, we deduce that the peripheries of both lungs are the important regions of the background of chest X-rays and we label segmentation maps of the lung peripheries for the positive class within the training set.

\begin{figure}[h]
\begin{center}
\centerline{\includegraphics[scale=0.18]{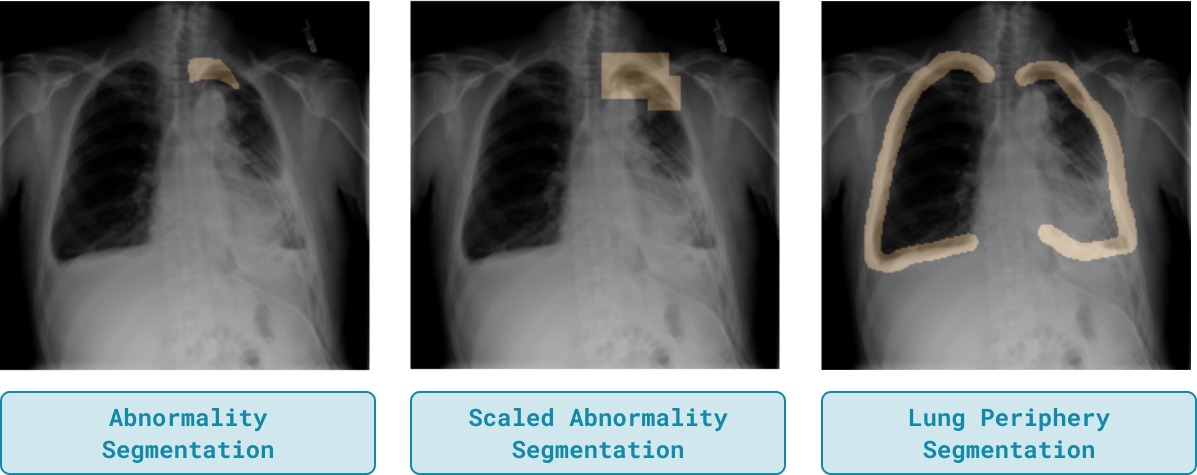}}
\caption{The highlighted area indicates the respective segmentation map. The lung periphery segmentation contains most of the abnormality along with additional background information.}
\label{segmentations}
\end{center}
\end{figure}

We hypothesize that incorporating these newly curated lung periphery segmentation maps which capture both the abnormality and relevant regions of the background will improve upon the original performance of ActDiff and RRR. In addition, we compare these new segmentation maps to a naive method of obtaining more of the image background by scaling the ground-truth segmentations to a smaller resolution so that they cover a larger area. \cref{segmentations} compares each of these types of segmentation maps for a given image. 


 We run experiments for both methods with each of the new segmentation maps. \cref{lung_periphery_results} reports results for ActDiff while results for RRR are in the Appendix. Lung periphery segmentations greatly increase the performance of ActDiff on the target sets, improving upon the ground-truth segmentations by an average of 6.32 points in AUROC. This highlights the effectiveness of including task-relevant regions outside the abnormality mask for OOD generalization.

\FloatBarrier
\FloatBarrier

\section{Scaling of Training Data}

With the incorporation of lung periphery segmentations, we exhibit improved OOD performance yet are still unable to consistently beat ERM. Given that both RRR and ActDiff work in a more specified setting of doing both binary classification and feature attribution to regions of interest, we investigate whether scaling up the number of training samples further improves the OOD performance of these methods. We hypothesize that to improve upon the initial failures of RRR and ActDiff, we require the combination of lung periphery segmentations and a sufficient sample size of training data. To verify this claim, we show that when training these methods on a larger dataset, ground-truth segmentations exhibit worse generalization performance in comparison to ERM. While on the other hand, lung periphery segmentations are able to improve upon ERM.

To construct a larger set, we scale our original source dataset from \cite{saab2021} to include the rest of the images in SIIM-ACR. We sample 8,540 examples for our training and validation split, reserving the remaining 2,135 images for a held out test set. We term this larger dataset CXR-P Full and label lung periphery segmentations for all positive example in the training split. Now training on CXR-P Full, \cref{full_results} demonstrates that with the use of lung-periphery segmentations we see gains on average of 1.35 points in AUROC over ERM across the OOD sets. In addition, we note that the lung-periphery segmentations also exhibit an average gain of 1.4 points in OOD performance over the use of ground-truth segmentations on CXR-P Full.

\FloatBarrier
\begin{table}[htb]
\caption{Test results when utilizing lung periphery segmentations and training on CXR-P Full. The Gain column specifies the difference in performance relative to using abnormality segmentations for the full set. Target datasets are underlined. The results are averaged over 3 random seeds with 95\% significance.}
\label{full_results}
\begin{center}
\begin{small}
\begin{sc}
\begin{tabular}{lcccr}
\toprule
Dataset & Method & AUROC & Gain \\
\midrule
\multirow{3}{*}{CXR-P Full} & ERM    & \textbf{90.6} $\pm$ 0.4& -  \\
& ActDiff & 88.3 $\pm$ 1.3& -1.3 \\
& RRR & 89.6 $\pm$ 0.9 & -1.3 &\\
\cline{0-3}
\multirow{3}{*}{\underline{MIMIC-CXR}} & ERM    & 78.3$\pm$ 1.1& -  \\
& ActDiff & \textbf{81.2}$\pm$ 0.6& +4.4 \\
& RRR & 80.4 $\pm$ 2.6 & +1.7 &\\
\cline{0-3}
\multirow{3}{*}{\underline{CheXpert}} & ERM    &  82.5$\pm$ 0.4& -  \\
& ActDiff & 82.6$\pm$ 1.5& +0.1 \\
& RRR & \textbf{82.8} $\pm$ 1.6 &  +1.0 &\\
\cline{0-3}
\multirow{3}{*}{\underline{CheXpert-age}} & ERM    & 73.2 $\pm$ 0.9 & -  \\
& ActDiff & 74.2$\pm$ 2.1& -0.3 \\
& RRR & \textbf{74.8} $\pm$ 1.4  & +1.9 & \\
\cline{0-3}
\multirow{3}{*}{\underline{ChestX-ray8}} & ERM    & 76.1 $\pm$ 0.8& -  \\
& ActDiff & \textbf{78.4}$\pm$ 0.6& +2.1 \\
& RRR & 76.6 $\pm$ 1.2  & +0.3 &\\
\bottomrule
\end{tabular}
\end{sc}
\end{small}
\end{center}
\end{table}
\FloatBarrier

\section{Conclusion}

By introducing a task-specific segmentation map that incorporates relevant regions of the background, we improved upon the initial failure in OOD performance of domain generalization methods that rely upon ground-truth segmentations. Utilizing these new segmentation maps along with a sufficient quantity of training data allows for these methods to beat ERM in OOD performance. Future work includes how to reduce reliance on a large number of training samples by utilizing concepts in weakly-, semi-, and self-supervised learning.








\bibliography{final_paper}
\bibliographystyle{icml2022}

\newpage
\appendix
\onecolumn
\section{Hyperparameter Tuning}
For all methods, we tune for learning rate (LR) in $\{ 1e-5, 1e-4, 1e-3 \}$ and weight decay (WD) in $\{0, 1e-4, 1e-3, 1e-2, 1e-1, 1 \}$. Additionally, for Actdiff we tune $\lambda_{actdiff}$ in $\{1e-5, 1e-4, 1e-3, 1e-2, 1e-1, 1 \}$ and for RRR we tune $\lambda_{rrr}$ in $\{1e-5, 1e-4, 1e-3, 1e-2, 1e-1, 1 \}$. The chosen hyperparameters for each method based on a validation performance is shown in \cref{hyperparmeters}

\FloatBarrier
\begin{table*}[h]
\caption{Best identfied hyperparameters per method.}
\label{hyperparmeters}
\begin{center}
\begin{small}
\begin{sc}
\begin{tabular}{lccccr}
\toprule
Method & LR & WD & Batch Size & $\lambda_{actdiff}$ & $\lambda_{rrr}$ \\
\midrule
ERM & 1e-4 & 0 & 16 & NA & NA \\ 
ActDiff with Abnormality Segmentations & 1e-4 & 0 & 16 & 1e-5 & NA \\
ActDiff with Lung Periphery Segmentations & 1e-4  & 0 &  16 & 1e-3 & NA  \\
RRR with Abnormality Segmentations & 1e-4 & 0 & 16 & NA & 1e-3  \\
RRR with Lung Periphery Segmentations & 1e-4 & 0 & 16 & NA & 1e-2   \\
\bottomrule
\end{tabular}
\end{sc}
\end{small}
\end{center}
\end{table*}

\section{Lung Periphery Segmentations vs Abnormality Segmentations for RRR}
We report the results for comparing lung periphery segmentations vs ground-truth segmentations for RRR in \cref{RRR}

\FloatBarrier
\begin{table*}[h]
\caption{Results for RRR when utilizing the different segmentation maps. The lung periphery segmentations consistently win across the target sets. Results are averaged over 10 random seeds with 95\% significance.}
\label{RRR}
\begin{center}
\begin{small}
\begin{sc}
\begin{tabular}{lcccccr}
\toprule
Segmentation Type & CXR-P & \underline{MIMIC-CXR} & \underline{CheXpert} & \underline{CheXpert-age} & \underline{ChestX-ray8} \\
\midrule
Abnormality & \textbf{83.5} $\pm$ 1.4 & 76.6$\pm$ 3.0 & 79.2 $\pm$ 2.0 & 71.6 $\pm$ 2.3 & \textbf{73.1} $\pm$ 1.6  \\ 
Lung Periphery & 83.2 $\pm$ 1.2 & \textbf{77.6} $ \pm$ 0.9 & \textbf{81.1} $\pm$ 2.8 & \textbf{73.0} $\pm$ 2.3 &  72.8 $\pm$ 1.6  \\
\bottomrule
\end{tabular}
\end{sc}
\end{small}
\end{center}
\end{table*}


\end{document}